\pdfoutput=1
\documentclass[11pt]{article}

\usepackage[final]{acl}

\usepackage{times}
\usepackage{latexsym}

\usepackage[T1]{fontenc}
\usepackage[utf8]{inputenc}

\usepackage{microtype}

\usepackage{inconsolata}

\usepackage{graphicx}
\usepackage{tabularx}

\title{Talk Before You Retrieve: Agent-Led Discussions for Better RAG in Medical QA}

\author{
\textbf{Xuanzhao Dong\textsuperscript{1*}},
\textbf{Wenhui Zhu\textsuperscript{1*}},
\textbf{Hao Wang\textsuperscript{2*}},
\textbf{Xiwen Chen\textsuperscript{2*}},\\
\textbf{Peijie Qiu\textsuperscript{3}},
\textbf{Rui Yin\textsuperscript{1}},
\textbf{Yi Su\textsuperscript{4}},
\textbf{Yalin Wang\textsuperscript{1}}\\
\textsuperscript{1}Arizona State University,
\textsuperscript{2}Clemson University,
\textsuperscript{3}Washington University in St.Louis\\
\textsuperscript{4}Banner Alzheimer’s Institute\\
\small{\textbf{Correspondence:} \href{mailto:email@domain.com}{xdong64@asu.edu}}
}

\begin{document}
\maketitle
\begin{abstract}
Medical question answering (QA) is a reasoning-intensive task that remains challenging for large language models (LLMs) due to hallucinations and outdated domain knowledge. Retrieval-Augmented Generation (RAG) provides a promising post-training solution by leveraging external knowledge. However, existing medical RAG systems suffer from two key limitations: \textbf{(1)} a lack of modeling for human-like reasoning behaviors during information retrieval, and \textbf{(2)} reliance on suboptimal medical corpora, which often results in the retrieval of irrelevant or noisy snippets.  To overcome these challenges, we propose \textit{Discuss-RAG}, a plug-and-play module designed to enhance the medical QA RAG system through collaborative agent-based reasoning. Our method introduces a summarizer agent that orchestrates a team of medical experts to emulate multi-turn brainstorming, thereby improving the relevance of retrieved content. Additionally, a decision-making agent evaluates the retrieved snippets before their final integration. Experimental results on four benchmark medical QA datasets show that \textit{Discuss-RAG} consistently outperforms MedRAG, especially significantly improving answer accuracy by up to 16.67\% on BioASQ and 12.20\% on PubMedQA.  The code is available at \url{https://github.com/LLM-VLM-GSL/Discuss-RAG}.
\end{abstract}

\def\thefootnote{*}\footnotetext{These authors contributed equally to this paper.}

\section{Introduction}
%%% llms got great ability in medical domain but got problems
Large Language Models (LLMs) have significantly advanced a wide range of medical tasks~\cite{singhal2023large,nori2023capabilities,kim2024mdagents}. However, their reliance on next-token prediction makes them susceptible to generating hallucinated responses~\cite{ji2023survey}. Additionally, once trained, LLMs operate with static parameters, meaning their internal knowledge remains fixed and cannot adapt to newly emerging research~\cite{zhang2023large}. As a result, LLMs face notable limitations in dynamic, reasoning-intensive tasks (e.g., medical question answering (QA)), where both up-to-date knowledge and complex logical inference are essential.
%%%%
\begin{figure}[t]
  \centering
  \includegraphics[width=0.45\textwidth]{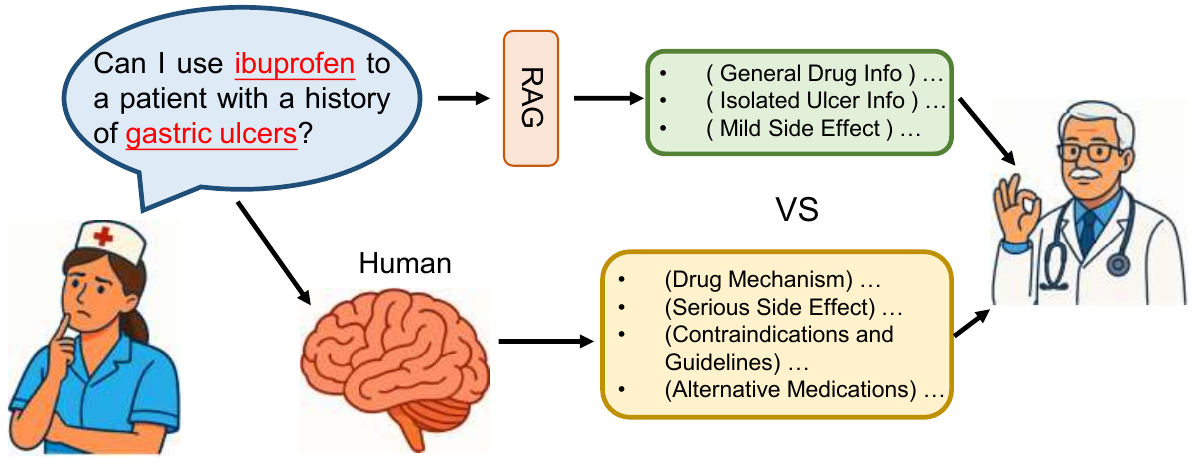}
  \caption{The illustration of difference between RAG and human for a medical query.}
  \label{fig:example}
\end{figure}
%%%%
% Despite these achievements, LLMs still face limitations in scenarios that require complex logical reasoning (e.g., medical question answering (QA)). Because they rely primarily on next-token prediction, LLMs are prone to generating plausible-sounding but factually inaccurate responses, known as hallucination. Additionally, once trained, LLMs are constrained by static parameters, which means their embedded knowledge does not evolve in real-time. This mismatch between static knowledge and the ever-changing landscape of medical understanding can hinder their reliability in clinical settings. These limitations underscore the challenges of applying LLMs in high-stakes domains such as diagnostic decision-making, and highlight the need for adaptable solutions. 
%% introduce RAG and focus on its problem
%%% emphasize bad corpora may raise problem, but usually in practice corpora is sub-optimal, hightlighting experiments in these case. maybe add main graph above and mention in this part.
Retrieval-Augmented Generation (RAG) has emerged as a promising approach to address the aforementioned limitations~\cite{borgeaud2022improving,guu2020retrieval,izacard2020leveraging}. By incorporating retrieved document snippets into the input prompt, RAG allows LLMs to generate responses that are grounded in up-to-date and trustworthy knowledge sources. Despite its success on several benchmarks, two concerns remain underexplored.
\begin{figure*}[t]
  \centering
  \includegraphics[width=0.9\textwidth]{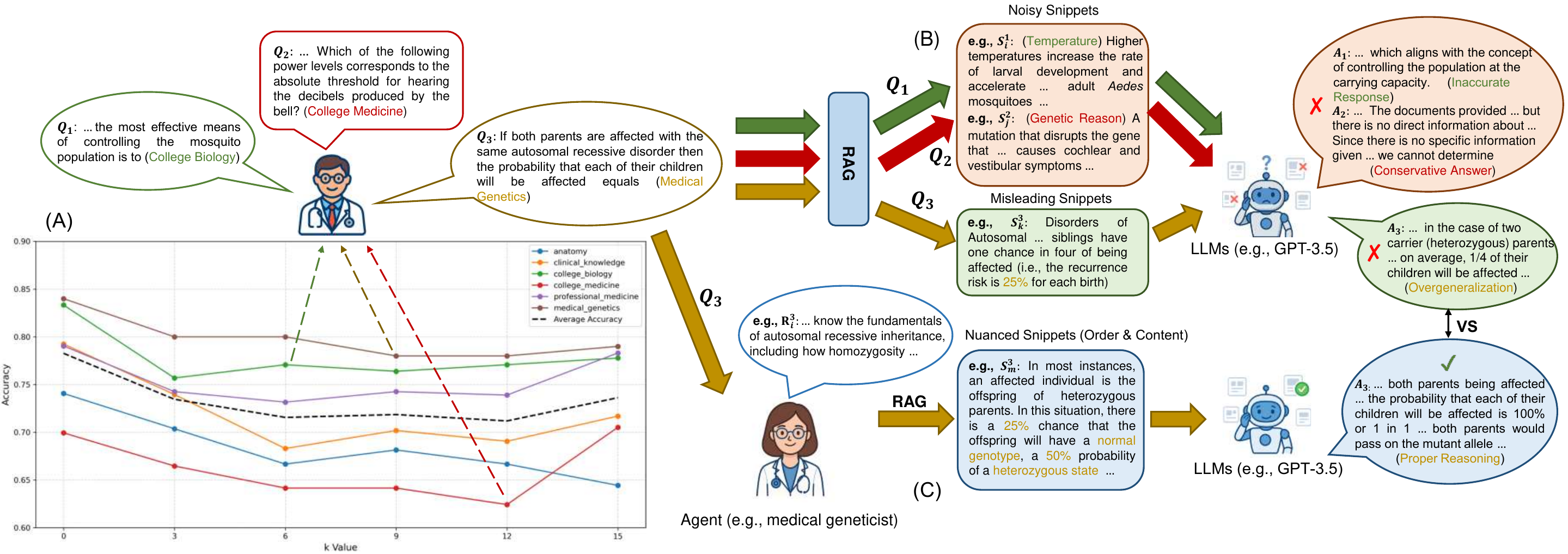}
  \caption{Preliminary experiments on the MMLU-Med benchmark. \textbf{(A)}. Accuracy trends as the number of retrieved documents k varies. Three representative questions ($Q_1,Q_2$ and $Q_3$) are selected to illustrate. \textbf{(B)}. Examples of retrieved snippets and the corresponding LLM (e.g., GPT-3.5) responses. \textbf{(C)}. Example of agent-led snippet selection and the resulting response for query $Q_3$. Additional details are discussed in Sec.~\ref{Sec:pre}.}
  \label{fig:pre-experiments}
\end{figure*}

% First, the lack of human-like information retrieval process, relying on statistical similarity metrics (e.g., cosine similarity) with questions to get relative document chunks, which leads to misleading retrieval. As shown in Fig., when nurses are asked whether ibuprofen can be administered to a patient with a history of gastric ulcers, they are more likely to recall relevant knowledge (e.g., contraindications of the medication) to make decision, rather than relying solely on surface-level semantic similarity to locate matching references. Second, the lack of post verification framework. Since medical queries and guidelines continually evolve, static corpora may leads to cautious or inaccurate responeses. As shown in Fig., we need a manager to make the final decision before going ahead.

First, current medical RAG systems lack a human-like information retrieval process. They typically rely on statistical similarity metrics (e.g., cosine similarity) between the query(e.g., questions) and document embeddings to retrieve relevant content~\cite{ke2024development}. This approach often fails to capture deeper contextual understanding, leading to the retrieval of superficially related but clinically irrelevant information. In contrast, as shown in Fig.~\ref{fig:example}, nurses in real-world clinical practice are more likely to recall and apply relevant clinical knowledge (e.g., drug contraindications) to guide decision-making, rather than relying solely on surface-level textual similarity. Second, existing systems often lack enough post-retrieval verification mechanisms~\cite{barnett2024seven,he2024retrieving}. Consequently, directly incorporating external knowledge may lead to overly cautious or outdated responses. In real-world settings, a judgmental role, such as a senior clinician reviewing a junior’s recommendation (Fig.~\ref{fig:example}), is often necessary to assess the correlation between retrieved context and context before a final decision is made.

To address these gaps between current medical RAG systems and real-world clinical decision-making processes, we proposed  \textit{Discuss-RAG}, an agent-led framework that enhances both the information retrieval and post-verification stages of medical RAG pipelines. Specifically, a summarizer agent collaborates with a team of specialized medical agents to generate progressively refined and context-rich background insights, which are incorporated into the retrieval process alongside the original query. Additionally, a decision-maker agent evaluates the relevance and coherence of the retrieved snippets and determines whether auxiliary components should be triggered. Notably, our framework is modular and can be seamlessly integrated into any existing training-free medical RAG pipeline. Experiments on four benchmark medical QA datasets demonstrate that \textit{Discuss-RAG} consistently improves response accuracy compared to baseline systems.

In summary, this paper makes the following key contributions: \textbf{\textit{(1)}}. We propose \textit{Discuss-RAG}, an agent-led RAG framework that simulates a human-like reference retrieval through multi-agent discussion and iterative summarization. \textbf{\textit{(2)}}. We introduce a post-retrieval verification agent that assesses the relevance and logical coherence of retrieved snippets before they are used in answer generation. \textbf{\textit{(3)}}. We conduct comprehensive experiments comparing \textit{Discuss-RAG} with standard RAG systems, demonstrating its effectiveness in improving both answer accuracy and snippet quality.

% \begin{itemize}
%     \item We propose \textit{Discuss-RAG}, an agent-led RAG framework that simulates a human-like reference retrieval through multi-agent discussion and iterative summarization.
%     \item Additionally, a post verification agent is prompted to evaluate the rationality of snippets for better response.
%     \item We conduct comprehensive experiments comparing \textit{Discuss-RAG} with standard RAG systems, demonstrating its effectiveness in improving both answer accuracy and snippet quality.
% \end{itemize}

%%%% Here is the plot for whole pipeline 
\begin{figure*}[t]
  \centering
  \includegraphics[width=0.9\textwidth]{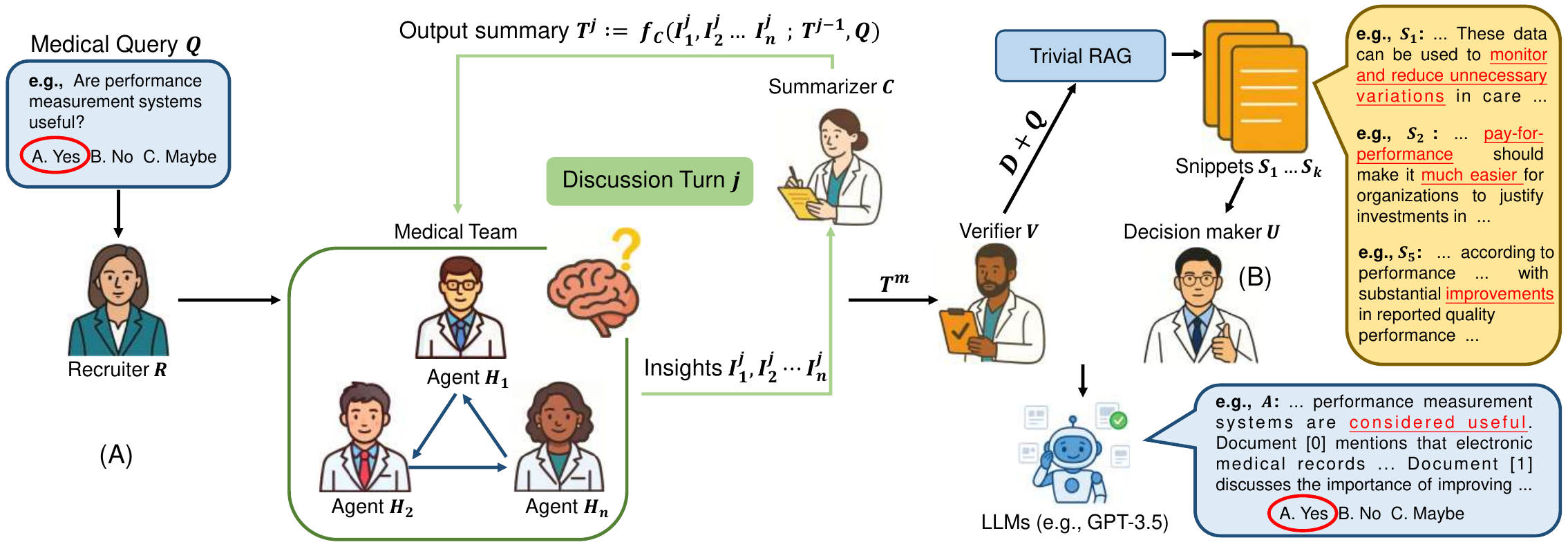}
  \caption{Illustration of the \textit{Discuss-RAG} pipeline. \textbf{(A)}. depicts the multi-turn brainstorming and summarization process. \textbf{(B)}. presents the agent-led post-retrieval verification module. A medical query, the corresponding snippets, and the LLM’s generated answer are used for illustration. Further details are provided in Sec~\ref{Sec:method}.
   }
  \label{fig:pipeline}
\end{figure*}
%%%%

\section{Preliminary}\label{Sec:pre}
% In practice, the number of retrieved documents k and the quality and relevance of the content retrieved influence the performance of RAG systems. While corpus quality is often assumed to be a variable, in many real-world medical settings the available corpus is fixed and inherently limited due to cost, and the dynamic nature of medical knowledge. To simulate this realistic constraint, we fix the corpus (i.e., textbooks) and examine how varying k influences LLM performance. As shown in Fig.~\ref{fig:pre-experiments}(A), changes in k result in fluctuating accuracy across six medical subjects, with performance generally remaining inferior or only comparable to that of chain-of-thought (CoT) prompting. This suggests that even with a static corpus, the number and selection of retrieved documents significantly impact downstream reasoning. To better understand this variation, we selected three representative questions ($Q_1,Q_2,Q_3$) at different k levels and performed a qualitative analysis of the retrieved snippets, which reveals two major issues contributing to suboptimal model behavior.
In our empirical experiments, we found that limitations hinder the performance of medical RAG systems in medical QA tasks. As shown in Fig.~\ref{fig:pre-experiments}(A), when the corpus is fixed (i.e., textbooks~\cite{jin2021disease}), varying the number of retrieved documents k results in fluctuating accuracy across six medical subjects. To better understand the influence of document selection, we selected three representative questions ($Q_1,Q_2,Q_3$) across different k values and subject domains. A qualitative analysis reveals factors contributing to suboptimal model behavior.

First, snippets selected based solely on dense vector similarity with the query often retrieve content that is conceptually related but task-irrelevant. These snippets introduce excessive background information that may confuse the LLM. As shown in Fig.~\ref{fig:pre-experiments}(B) for $Q_1$, high-scoring snippets focus on environmental factors such as climate and temperature in relation to mosquitoes, rather than addressing strategies for population control. This misalignment leads to noisy inputs, resulting in either inaccurate or overly cautious responses, as seen in $Q_2$. Second, even factually correct snippets can mislead the model. In the case of $Q_3$, retrieved snippets emphasize the $25\%$ probability associated with autosomal inheritance, prompting the LLM to overgeneralize from heterozygous to homozygous cases. These findings further suggest that directly using retrieved snippets without verification can lead to reasoning errors.

To further examine the limitations of hard similarity-based retrieval, we conducted an exploratory experiment using the same query ($Q_3$). As shown in Fig.~\ref{fig:pre-experiments}(C), we prompted a domain-specific agent (i.e., a medical geneticist) to identify the essential knowledge required to answer the question (mimicking the behavior of nurses, as illustrated in Fig.~\ref{fig:example}). When we used the agent’s response, in conjunction with the original query, to guide retrieval, the resulting snippets were both more topically relevant and better organized. Under this setting, the LLM successfully distinguished between carriers and affected individuals and generated a well-reasoned response. 

These findings motivate two key directions for better medical RAG: \textbf{\textit{(1)}}. While a single role-based agent can benefit retrieval quality, can a multi-agent setup, engaging diverse medical expertise in an iterative, self-refining discussion, yield a more comprehensive and contextually rich background? \textbf{\textit{(2)}}. Given that structured agent involvement benefits retrieval, can a similar structure be extended to the response stage? To address these questions, we propose an agent-led RAG paradigm, the details of which are presented in the following section.

\section{Methodology}\label{Sec:method}
% Motivated by the findings from our preliminary study, we propose an agent-led retrieval-augmented generation (RAG) pipeline designed to enhance information retrieval. Our proposed method, termed Discuss-RAG, consists of two key components: (1) a human-like multi-turn discussion and summarization module (Fig.~\ref{fig:pipeline}(A)), and (2) a post-processing verification stage (Fig.~\ref{fig:pipeline}(B)). These components are inspired by the cognitive processes of collaborative brainstorming and subsequent validation, respectively.

% Based on insights discussed in Sec.~\ref{Sec:pre}, we introduce \textit{Discuss-RAG}, a modular, training-free framwork compose: (1). a human-like multi-turn discussion and summarization module. (2). a post-retrieval verification module.
Our method contain two components: (1) A human-like multi-turn discussion and summarization module. (2) A post-retrieval verification module.

% \subsection{Multi-turn Discussion and summarization}\label{Sec:discussion}

\noindent\textbf{Multi-turn Discussion and summarization}. This module simulates a collaborative brainstorming process between a team of medical experts and a summarizer (acting as a moderator). Specifically, given a medical query $Q$, a recruiter agent $R$ assembles a team of medical domain experts $H_i$ ( for $i$ in $1,2\dots n$), each contributing their domain-specific perspectives $I_i^j$ at turn $j$ ( for $j$ in $0,1\dots m$). A summarizer agent $C$ is then prompted to extract key medical knowledge, background concepts, and reasoning steps from these inputs to generate a concise summary $T^j$. This iterative process is formally denoted as:
\begin{equation}
    T^j := f_C(I_1^j, I_2^j, \ldots, I_n^j \,;\, T^{j-1}, Q)
\end{equation}
Here $f_C(\cdot)$ denotes the summarization process performed by agent $C$, and $T^j$ reflects the progressively refined understanding of the query, based on the current reflection $I_i^j$, previous summary $T^{j-1}$ and the original query $Q$ ( with $T^0$ initialized as an empty summary). After the discussion concludes, a verifier agent $V$ is introduced to evaluate the consistency and sufficiency of the final summary $T^m$. The verifier produces a distilled, verification-passed summary $D$, which is subsequently used for snippet retrieval, together with the original query $Q$. 
%%%% Here is the plot for result illustration
\begin{figure*}[t]
  \centering
  \includegraphics[width=0.9\textwidth]{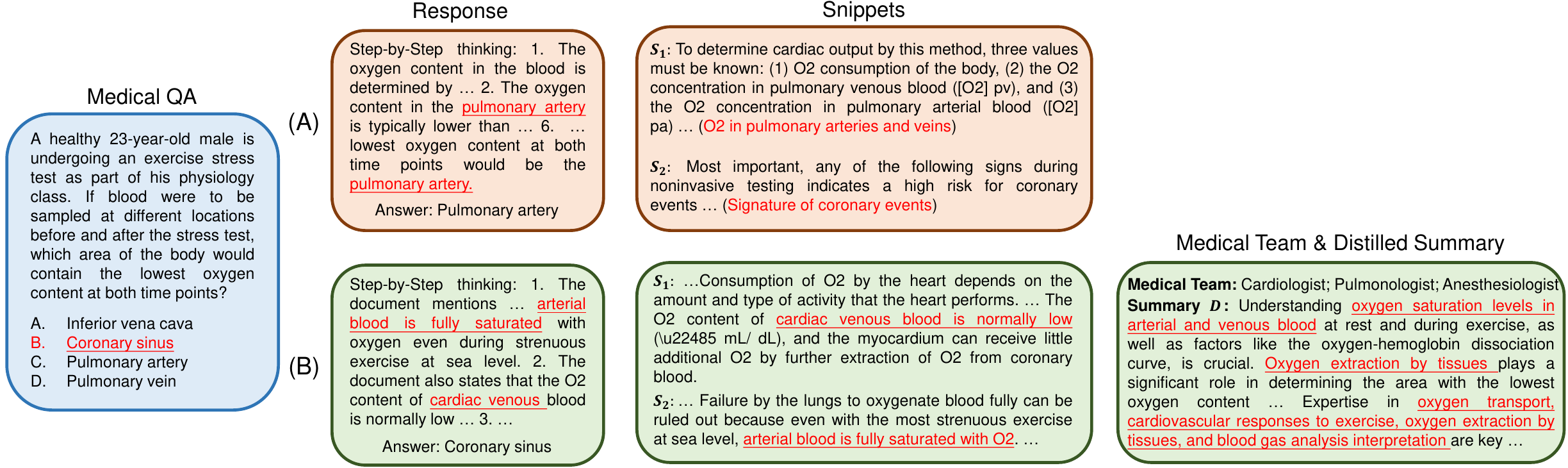}
  \caption{Example from the MedQA-US benchmark comparing MedRAG \textbf{(A)} and Discuss-RAG \textbf{(B)}. Answers and key phrases are highlighted in red.}
  \label{fig:result}
\end{figure*}

As shown in Fig.~\ref{fig:pipeline}(A), the recruiter $R$ recruits a team consisting of three specialized agents (e.g., a health care quality specialist, a hospital administrator, and a health economist), who collaborate with the summarizer $C$ to share their insights for the performance measurement system. The conversation terminates either when the maximum number of discussion rounds $m$ is reached, or when all agents decline to contribute further. Notably, all agents in this module are explicitly instructed not to answer the original query or infer a final conclusion. This design ensures that the process remains focused on context construction for retrieval, rather than direct answer generation.

\noindent\textbf{Post-retrieval Verification}. This module leverages structured agent reasoning to mitigate the adverse effects of suboptimal retrieval. Specifically, given the distilled summary $D$ and the medical query $Q$, a specialized decision-maker agent $U$ is introduced to evaluate the top-$k$ document chunks $S_i$ retrieved by the underlying retrieval algorithm. If $U$ returns a negative judgment, an alternative retrieval strategy is triggered (e.g., a CoT-based prompt~\cite{wei2022chain} is used as a fallback in our implementation). Otherwise, the accepted snippets are incorporated into the context prompt for answer generation. As shown in Fig.~\ref{fig:pipeline}(B), the verified snippets tend to be closely aligned with the intended focus of the query. In the shown example, the selected evidence explicitly highlights the effect (marked in red) of performance measurement systems, providing grounded support for a more accurate and contextually appropriate response.

It is important to emphasize that both components of our proposed pipeline are designed as plug-and-play enhancements for any training-free RAG system. The multi-turn brainstorming process enriches the information fed to the retriever, while the post-retrieval verification module dynamically filters and validates the retrieved content, providing improved reliability and quality in the answer.

\section{Experiments}
\textbf{Experimental details}. We evaluated the retrieval ability of our method on four medical QA benchmark datasets: MMLU-Med~\cite{hendrycks2020measuring}, MedQA-US~\cite{jin2021disease}, BioASQ~\cite{tsatsaronis2015overview}, and PubMedQA~\cite{jin2019pubmedqa}. MedRAG~\cite{xiong2024benchmarking}. To ensure a fair comparison, we employed the same medical textbooks~\cite{jin2021disease} as the corpus and MedCPT~\cite{jin2023medcpt} as the retriever with MedRAG. For LLM, GPT-3.5 (i.e., gpt-3.5-turbo-0125~\cite{openai2024gpt35turbo}) was selected. 
% For other necessary parameters, we chose $n=3$, $k=9$, and $m=2$.
%%% insert table about benchmark info %%%%
% \begin{table}[h]
%     \centering
%     \caption{Benchmark Dataset Statistics. \#Cho indicates the number of answer options.  }
%     \label{tab: benchmark}
%     \begin{tabular}{|c|c|c|c|}
%         \hline
%         \textbf{Dataset} & \textbf{Size} & \textbf{\#Cho} & \textbf{Type} \\ 
%         \hline
%         MMLU-Med & 1089 & 4 & Examination \\
%         MedQA-US & 1273 & 4 & Examination \\
%         BioASQ & 720 & 2 & Literature \\
%         PubMedQA & 500 & 3 & Literature \\
%         \hline
%     \end{tabular}
% \end{table}

% \begin{table}[h]
%     \centering
%     \caption{Benchmark dataset results. Answer accuracy was used as the evaluation metric.}
%     \label{tab:results}
%     \small
%     \resizebox{\columnwidth}{!}{
%     \begin{tabular}{|c|c|c|}
%         \hline
%         \textbf{Dataset (Task)} & \textbf{MedRAG} &\textbf{MedRAG + Discuss-RAG} \\ 
%         \hline
%         MMLU-Med & 0.7153 & 0.7723 \\
%         MedQA-US & 0.6245 & 0.6685 \\
%         BioASQ & 0.5861 & 0.7528 \\
%         PubMedQA & 0.3560 & 0.4780 \\
%         \hline
%     \end{tabular} }
% \end{table}

\begin{table}[h]
    \centering
    \caption{Benchmark dataset results. Answer accuracy was used as the evaluation metric.}
    \label{tab:results}
    \small
    \resizebox{\columnwidth}{!}{
    \begin{tabular}{|c|c|c|c|}
        \hline
        \textbf{Dataset} & \textbf{MedRAG} & \textbf{+ Discuss-RAG} & \textbf{$\Delta$} \\ 
        \hline
        MMLU-Med & 71.53\% & 77.23\% & \textcolor{red}{+5.70\%} \\
        \hline
        MedQA-US & 62.45\% & 66.85\% & \textcolor{red}{+4.40\%} \\
        \hline
        BioASQ & 58.61\% & 75.28\% & \textcolor{red}{+16.67\%} \\
        \hline
        PubMedQA & 35.60\% & 47.80\% & \textcolor{red}{+12.20\%} \\
        \hline
    \end{tabular} }
\end{table}

\noindent\textbf{Experimental results and analysis}. Leveraging the multi-turn discussion and post-retrieval verification modules, \textit{Discuss-RAG} enriches the background information available and mitigates the impact of suboptimal retrieval. As shown in Tab.~\ref{tab:results}, integrating our method consistently improves MedRAG performance across all four benchmarks, especially achieving gains of up to 16.67\% on the BioASQ dataset and 12.20\% on PubMedQA.
Further, as illustrated in Fig.~\ref{fig:result}, for the same query, the top-2 snippets retrieved by \textit{Disscuss-RAG} provide more grounded and factual support for correctly answering the question. Specifically, snippets $S_1$ explicitly mention the low oxygen ($O_2$) content in cardiac venous blood, while snippets $S_2$ support the reasoning process from a contrasting perspective. Additionally, the final distilled summary $D$ generated by the medical team highlights the essential knowledge required to focus the retrieval process, leading to more reliable and contextually appropriate evidence selection.

\section{Conclusion}
% In this work, we propose \textit{Discuss-RAG}, an agent-led, plug-and-play framework that could be integrated into any training-free RAG pipeline to enhance the response accuracy of LLMs in medical QA. To address the challenge of irrelevant retrieval caused by purely semantic similarity-based search with query, we design a multi-turn discussion and summarization module that facilitates context-rich and self-refined document retrieval. To mitigate the negative impact of suboptimal snippets, we introduce a post-retrieval verification agent that makes a final judgment before response generation. Experiments on four medical QA benchmarks dataset demonstrate that \textit{Discuss-RAG} improves both response accuracy and snippets quality.
% In this work, we propose \textit{Discuss-RAG}, an agent-led framework used to enhance the response accuracy of LLMs in medical QA. Specifically, a multi-turn discussion and summarization module is designed to facilitate context-rich and self-refined document retrieval, and a post-retrieval verification agent is prompted to make final judgment. Experiments on four medical QA benchmarks dataset demonstrate that \textit{Discuss-RAG} improves both response accuracy and snippets quality
In this work, we propose \textit{Discuss-RAG}, an agent-led framework designed to enhance the response accuracy of LLMs in medical QA. Specifically, we introduce a multi-turn discussion and summarization module to facilitate context-rich and self-refined document retrieval, and a post-retrieval verification agent to make the final judgment on the retrieved content. Experiments conducted on four medical QA benchmark datasets demonstrate that \textit{Discuss-RAG} consistently improves both response accuracy and snippet quality.

\section{Limitation}
We acknowledge that \textit{Discuss-RAG} is hindered by two primary limitations. \textbf{(1)}. Limited interaction among team members. The specialized medical agents $H_i$ do not communicate directly with one another, but interact through the summary from the previous round. Direct peer-to-peer interaction may facilitate deeper and more dynamic reasoning. \textbf{(2)}. Increased computational overhead. Our framework involves prompting multiple LLM-based agents, each requiring careful instruction design to perform their respective roles effectively. This introduces additional computational and engineering costs.

\bibliography{custom}

% \appendix

% This is an appendix.

\end{document}